\documentclass[aos,preprint]{imsart}
\setattribute{journal}{name}{} 

\RequirePackage[OT1]{fontenc}
\RequirePackage{amsthm,amsmath,amssymb}

\RequirePackage{natbib}
\setcitestyle{round,authoryear,citesep={;},aysep={,},yysep={;}}

\renewcommand{\cite}[1]{\citep{#1}}

\RequirePackage[colorlinks,citecolor=blue,urlcolor=blue]{hyperref}

\usepackage{graphicx} % more modern

\usepackage{macros}

\usepackage{caption}
\usepackage{subcaption}

\usepackage{verbatim}
\usepackage{gensymb}

\usepackage[top=1in, bottom=1in, left=1in, right=1in]{geometry}

\usepackage{macros}
\usepackage{savesym}
\savesymbol{captionbox}
\usepackage[font={small}]{caption}
\DeclareCaptionType{copyrightbox}
\usepackage{subcaption}
\restoresymbol{CAP}{captionbox}

\usepackage{amsmath,amssymb}
\usepackage{verbatim}
\usepackage{gensymb}
\usepackage{multirow}

\begin{document}

\begin{frontmatter}
\title{Dependent Multinomial Models Made Easy: 
\\Stick Breaking with the P\'{o}lya-Gamma Augmentation}
\runtitle{Dependent Multinomial Models Made Easy}

\begin{aug}
\author{\fnms{Scott W.} \snm{Linderman},\thanksref{t1}\ead[label=e1]{swl@seas.harvard.edu}}
\author{\fnms{Matthew J.} \snm{Johnson},\thanksref{t1}\ead[label=e2]{mattjj@csail.mit.edu}}
\and
\author{\fnms{Ryan P.} \snm{Adams}\ead[label=e3]{rpa@seas.harvard.edu}}
\affiliation{Harvard University}

\thankstext{t1}{These authors contributed equally.}

\runauthor{S. W. Linderman, M. J. Johnson, and R. P. Adams}
\end{aug}

%\maketitle
\begin{abstract}
Many practical modeling problems involve discrete data that are best represented as draws from multinomial or categorical distributions. For example, nucleotides in a 
DNA sequence, children's names in a given state and year, and text documents are all commonly modeled with multinomial distributions.  In all of these cases, we expect 
some form of dependency between the draws: the nucleotide at one position in the DNA strand may depend on the preceding nucleotides, children's names are highly 
correlated from year to year, and topics in text may be correlated and dynamic.  These dependencies are not naturally captured by the typical Dirichlet-multinomial 
formulation.  Here, we leverage a logistic stick-breaking representation and recent innovations in P\'{o}lya-gamma augmentation to reformulate the multinomial distribution 
in terms of latent variables with jointly Gaussian likelihoods, enabling us to take advantage of a host of Bayesian inference techniques for Gaussian models with minimal overhead.
\end{abstract}
\end{frontmatter}

\section{Introduction}

It is often desirable to model discrete data in terms of continuous latent structure.
In applications involving text corpora, discrete-valued time series, or polling and
purchasing decisions, we may want to learn correlations or spatiotemporal
dynamics, and leverage these learned structures to improve inferences and
predictions.
However, adding these continuous latent dependence structures often comes at the
cost of significantly complicating inference: such models may require
specialized, one-off inference algorithms, such as a nonconjugate variational
optimization, or they may only admit very general inference tools like particle
MCMC \cite{andrieu2010particle} or elliptical slice sampling \cite{murray-adams-mackay-2010a}, which can be inefficient and difficult to scale.
Developing, extending, and applying these models has remained a challenge.

In this paper we aim to provide a class of such models that are easy and
efficient.
We develop models for categorical and multinomial data in which dependencies
among the multinomial parameters are modeled via latent Gaussian distributions
or Gaussian processes, and we show that this flexible class of models admits a
simple auxiliary variable method that makes inference easy, fast, and modular.
This construction not only makes these models simple to develop and apply, but
also allows the resulting inference methods to use off-the-shelf
algorithms and software for Gaussian processes and linear Gaussian dynamical
systems.

The paper is organized as follows.
After providing background material and defining our general models and
inference methods, we demonstrate the utility of this class of models by applying it
to three domains as case studies.
First, we develop a correlated topic model for text corpora.
Second, we study an application to modeling the spatial and temporal patterns
in birth names given only sparse data.
Finally, we provide a new continuous state-space model for discrete-valued
sequences, including text and human DNA.
In each case, given our model construction and auxiliary variable method,
inference algorithms are easy to develop and very effective in experiments.
We conclude with comments on the new kinds of models these
methods may enable.

Code to use these models and reproduce all the figures is available at
\url{https://github.com/HIPS/pgmult}.

\section{Modeling correlations in multinomial parameters}
\label{sec:prelim}

In this section, we discuss an auxiliary variable scheme that allows
multinomial observations to appear as Gaussian likelihoods within a larger
probabilistic model.  The key trick discussed in the proceeding sections is to
introduce P\'{o}lya-gamma random variables into the joint distribution over
data and parameters in such a way that the resulting marginal leaves the
original model intact.

\subsection{P\'{o}lya-gamma augmentation}
The integral identity at the heart of the P\'{o}lya-gamma augmentation scheme \cite{polson2013bayesian} is
\begin{equation}
\label{eq:pg_identity}
\frac{(e^{\psi})^a}{(1+e^{\psi})^b} = 2^{-b} e^{\kappa \psi} \int_{0}^{\infty} e^{-\omega \psi^2 /2} p(\omega \given b, 0) \, \mathrm{d}\omega,
\end{equation}
where~${\kappa=a-b/2}$ and~$p(\omega\given b, 0)$ is the density of the P\'{o}lya-gamma
distribution~${\distPolyaGamma(b, 0)}$, which does not depend on $\psi$.
% Notice that for fixed $\omega$ the right-hand side is log-quadratic in $\psi$,
% while the left-hand side resembles a logistic link function common in many
% likelihood functions of interest.
% This structure, along with the fact that P\'{o}lya-gamma variates can be simulated
% efficiently \cite{}, leads to the augmentation scheme \cite{}.
Consider a likelihood function of the form
\begin{equation}
  \label{eq:logit_likelihood}
  p(x \given \psi) = c(x) \frac{(e^\psi)^{a(x)}}{(1+e^\psi)^{b(x)}}
\end{equation}
for some functions $a$, $b$, and $c$.
Such likelihoods arise, e.g., in logistic regression and in binomial
and negative binomial regression \cite{polson2013bayesian}.
Using~\eqref{eq:pg_identity} along with a prior~$p(\psi)$, we can write the joint density of $(\psi, x)$ as
\begin{equation}
  \label{eq:pg_joint}
  p(\psi, x) = p(\psi) \, c(x) \frac{(e^\psi)^{a(x)}}{(1+e^\psi)^{b(x)}} = \int_0^\infty
  p(\psi) \, c(x) \, 2^{-b(x)} e^{\kappa(x) \psi} e^{-\omega \psi^2/2} p(\omega \given b(x), 0) \; \mathrm{d}\omega.
\end{equation}
The integrand of~\eqref{eq:pg_joint} defines a joint density on $(\psi, x,
\omega)$ which admits $p(\psi, x)$ as a marginal density.
Conditioned on these auxiliary variables $\omega$, we have
\begin{equation}
  p(\psi \given x, \omega) \propto p(\psi) e^{\kappa(x) \psi} e^{- \omega \psi^2/2}
\end{equation}
which is Gaussian when $p(\psi)$ is Gaussian.
Furthermore, by the exponential tilting property of the P\'{o}lya-gamma
distribution, we have ${\omega \given \psi, x \sim \distPolyaGamma(b(x), \psi)}$.
Thus the identity~\eqref{eq:pg_identity} gives rise to a conditionally conjugate
augmentation scheme for Gaussian priors and likelihoods of the
form~\eqref{eq:logit_likelihood}. 
%Efficient algorithms and \texttt{C} implementations have been developed for sampling P\'{o}lya-gamma random variates \cite{windle2014sampling} and we have wrapped these implementations with a Python wrapper: \url{https://github.com/slinderman/pypolyagamma}.

This augmentation scheme has been used to develop Gibbs sampling and
variational inference algorithms for Bernoulli, binomial \cite{polson2013bayesian}, and
negative binomial regression models \cite{zhou2012lognormal} with logit link functions.
In this paper we extend it to the multinomial distribution.

\subsection{A P\'{o}lya-gamma augmentation for the multinomial distribution}
\label{sec:pg-aug}
\begin{figure}
\centering
% Top row: \psi's
  \begin{subfigure}[t]{.32\textwidth}
    \centering
    \vskip 0pt
    \includegraphics[width=\textwidth]{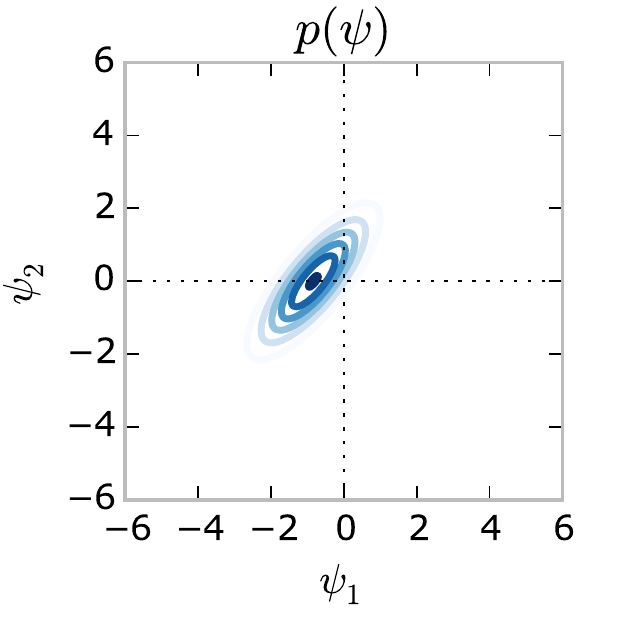}
    \label{fig:1b_psi}
  \end{subfigure}
  ~
  \begin{subfigure}[t]{.32\textwidth}
    \centering
    \vskip 0pt
    \includegraphics[width=\textwidth]{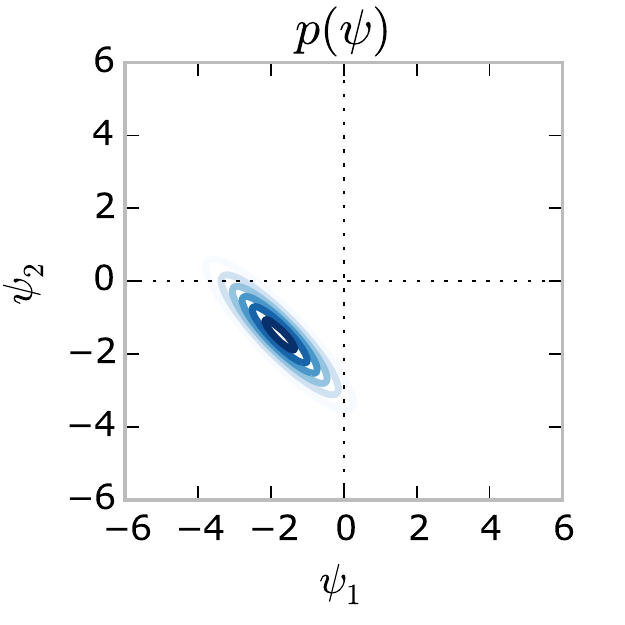}
    \label{fig:1b_psi}
  \end{subfigure}
  ~
    \begin{subfigure}[t]{.32\textwidth}
    \centering
    \vskip 0pt
    \includegraphics[width=\textwidth]{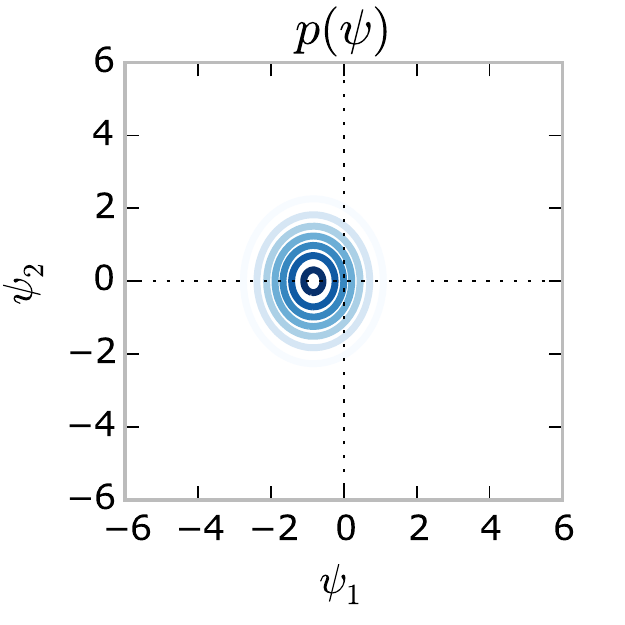}
    \label{fig:1b_psi}
  \end{subfigure}
  \\
 % Bottom row: \pi's
 \vspace{-2em}
  \begin{subfigure}[t]{.32\textwidth}
    \centering
    \vskip 0pt
    \includegraphics[width=\textwidth]{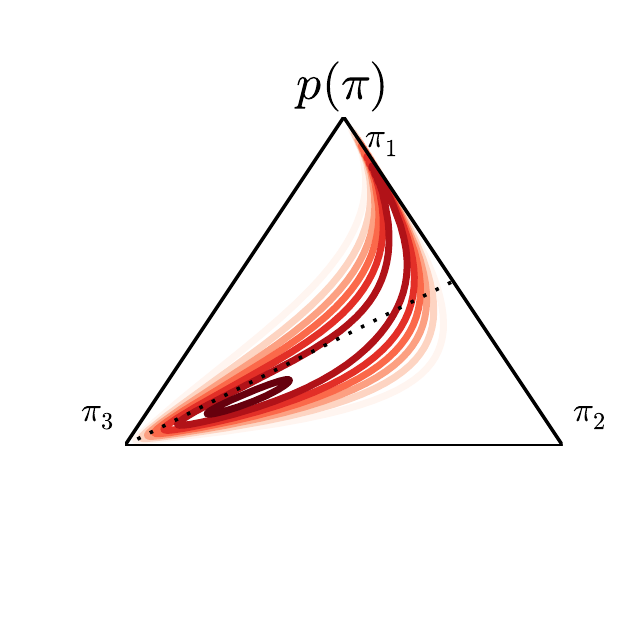}
    \label{fig:1b_psi}
  \end{subfigure}
  ~
  \begin{subfigure}[t]{.32\textwidth}
    \centering
    \vskip 0pt
    \includegraphics[width=\textwidth]{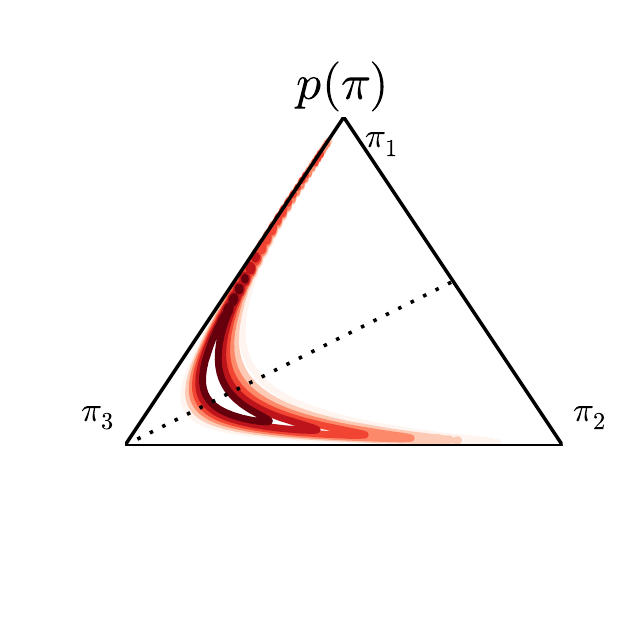}
    \label{fig:1b_psi}
  \end{subfigure}
  ~
    \begin{subfigure}[t]{.32\textwidth}
    \centering
    \vskip 0pt
    \includegraphics[width=\textwidth]{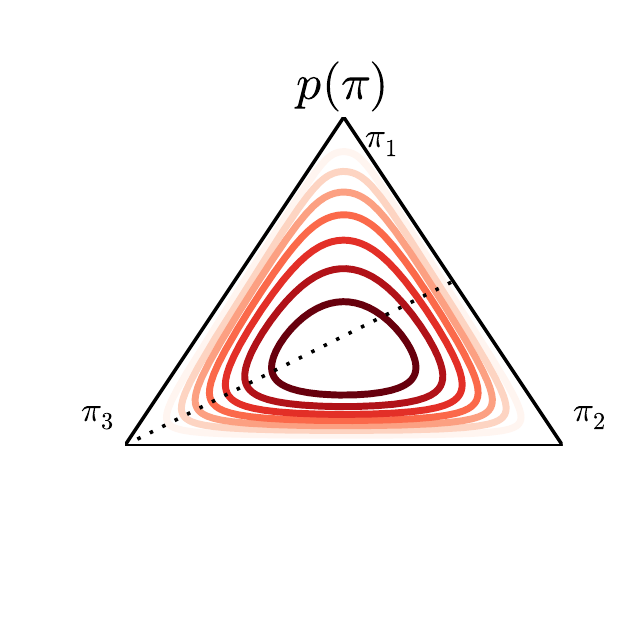}
    \label{fig:1b_psi}
  \end{subfigure}
 
    \vspace{-4em}
  \caption{Correlated 2D Gaussian priors on~$\bpsi$ and their implied densities on~$\pisb(\bpsi)$. See text for details. }
  \vspace{-1em}
\label{fig:pgdensity}
\end{figure}

To develop a P\'{o}lya-gamma augmentation for the multinomial, we first rewrite the
$K$-dimensional multinomial density recursively in terms of ${K-1}$ binomial
densities:
\begin{gather}
\distMultinomial(\bx \given N, \bpi) = \prod_{k=1}^{K-1} \distBinomial( x_k \given N_k, \widetilde{\pi}_k), \\
N_k = N - \sum_{j < k} x_j, \quad
\widetilde{\pi}_k = \frac{\pi_k}{1-\sum_{j < k} \pi_j}, \quad
k=2,3,\ldots,K,
\end{gather}
where $N_1=N=\sum_k x_k$ and $\widetilde{\pi}_1= \pi_1$.
For convenience, we define~${N(\bx)\equiv  \left[N_1, \ldots, N_{K-1} \right]}$.
This decomposition of the multinomial density is a ``stick-breaking''
representation where each~$\widetilde{\pi}_k$ represents the fraction of
the remaining probability mass assigned to the~$k$-th component.
We let~$\widetilde{\pi}_k = \sigma(\psi_k)$ and define the 
function,~${\pisb:\reals^{K-1} \to [0,1]^{K}}$,  
which maps a vector~$\bpsi$ to a normalized probability vector~$\bpi$. 

%Next, we rewrite the density into the form required by~\eqref{eq:pg_identity}
%by writing~$\bpi = \pisb(\bpsi)$ for~${\bpsi \in \reals^{K-1}}$ and
%substituting:
Next, we rewrite the density into the form required by~\eqref{eq:pg_identity}
by substituting $\sigma(\psi_k)$ for $\widetilde{\pi}_k$:
\begin{align}
\distMultinomial(\bx \given N, \bpsi)
= \prod_{k=1}^{K-1} \distBinomial(x_k \given N_k, \sigma(\psi_k))
&= \prod_{k=1}^{K-1} { N_k \choose x_k } \sigma(\psi_k)^{x_k} (1-\sigma(\psi_k))^{N_k - x_k} \\
&= \prod_{k=1}^{K-1} { N_k \choose x_k } \frac{(e^{\psi_k})^{x_k}}{(1+e^{\psi_k})^{N_k}}.
\end{align}
Choosing ${a_k(x)=x_k}$ and ${b_k(x)=N_k}$ for each ${k=1,2,\ldots,K-1}$, we can
then introduce P\'{o}lya-gamma auxiliary variables $\omega_k$ corresponding to each
coordinate~$\psi_k$;
dropping terms that do not depend on~$\bpsi$ and completing the square yields
\begin{align}
p(\bx, \bomega \given \bpsi) \propto \prod_{k=1}^{K-1} e^{(x_k-N_k/2)\psi_k -\omega_k \psi_k^2/2}
%&\propto \prod_{k=1}^{K-1} \distNormal \left( \psi_k \,\bigg|\, \frac{x_k-N_k/2}{\omega_k}, \frac{1}{\omega_k} \right) 
&\propto \distNormal \left(\bpsi \,\bigg|\, \bOmega^{-1} \kappa(\bx),\, \bOmega^{-1} \right),
\end{align} 
where~${\bOmega \equiv \text{diag}(\bomega)}$ and~${\kappa(\bx) \equiv \bx - N(\bx)/2}$.
That is, conditioned on~$\bomega$, the likelihood of~$\bpsi$ under the
augmented multinomial  model is proportional to a diagonal Gaussian distribution.

Figure~\ref{fig:pgdensity} illustrates how a variety of Gaussian densities map
to probability densities on the simplex. Correlated Gaussians (left) put most
probability mass near the~$\pi_1=\pi_2$ axis of the simplex, and
anti-correlated Gaussians (center) put mass along the sides where~$\pi_1$ is
large when~$\pi_2$ is small and vice-versa.
Finally, a nearly spherical Gaussian approximates a symmetric Dirichlet
distribution. Appendix~A derives a closed-form expression for the density
on~$\bpi$ implied by a Gaussian distribution on~$\bpsi$, as well an expression
for a diagonal Guassian distribution that best approximates, in a
moment-matching sense, a Dirichlet distribution on~$\bpi$.
 
% Briefly describe the logistic normal transformation 
\subsection{Alternative models}
This stick breaking transformation has been explored in the previous work
of~\citet{ren2011logistic} and~\citet{khan2012stick}, but has not been
connected with the P\'{o}lya-gamma augmentation. The multinomial probit and
logistic normal methods are most commonly used, and we describe them here.

% Discuss the multinomial probit model
The multinomial probit model \cite{albert1993bayesian} applies to categorical regrssion, and is based upon the following auxiliary variable model:~${z_k = \Phi(\psi_k + \epsilon_k)}$, where~$\Phi$ is the probit function and~$\epsilon_k \sim \distNormal(0,1)$. Given these auxiliary variables,~${x_k = 1}$ if~${z_k > z_j\, \forall j \neq k }$, and~${x_k=0}$ otherwise. 
%This has been extended to categorical Gaussian process regression \cite{girolami2006variational} using a variational Bayesian inference algorithm. 
This approach has primarily focused on categorical modeling.

A common method of modeling correlated multinomial parameters, to which we
directly compare in the following sections, is based on the ``softmax'' or
multi-class logistic function,~${\pi_k = e^{\psi_k} / \sum_{j=1}^K
e^{\psi_j}}$.
Let~$\piln: \reals^K \to [0,1]^K$ denote the joint transformation.
This has found application in multiclass regression \cite{holmes2006bayesian} and is the common approach to correlated topic modeling \cite{blei2006correlated}. 
However, leveraging this transformation in conjunction with Gaussian models for~$\bpsi$ is challenging 
due to the lack of conjugacy, and previous work has relied upon variational approximations 
to tackle the hard inference problem \cite{blei2006correlated}.

Unlike the logistic normal and multinomial probit, the stick-breaking
transformation we employ is asymmetric. Our illustrations in
Figure~\ref{fig:pgdensity}  and the discussion in Appendix A show that this
lack of symmetry does not impair the representational capacity of the model.

% Correlated Topic Models applied to newsgroup data
\section{Correlated topic models}

\begin{figure}
\centering
%  \begin{subfigure}[t]{.32\textwidth}
%    \centering
%    \vskip 0pt
%    \includegraphics[width=\textwidth]{perp_vs_num_topics}
%    \label{fig:perp_vs_num_topics}
%  \end{subfigure}
\begin{subfigure}[t]{1.75in}
    \centering
    \vskip -7pt
    \includegraphics[width=1.75in]{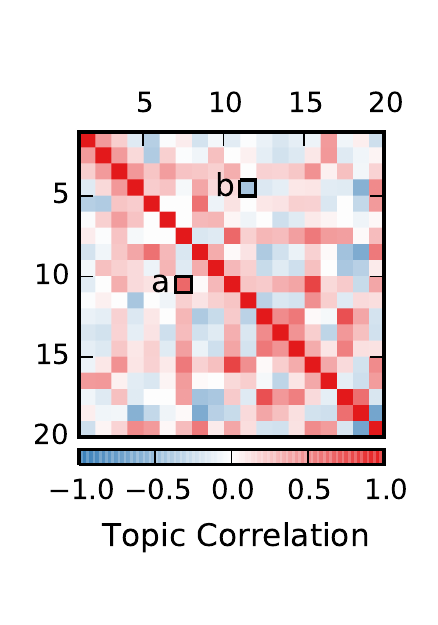}
    \label{fig:ctm_corr_matrix}
  \end{subfigure}
  ~
  \begin{subfigure}[t]{1.75in}
    \centering
    \vskip 0pt
    \includegraphics[width=\textwidth]{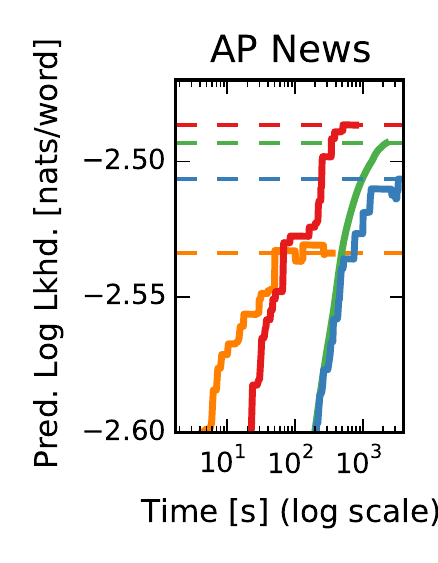}
    \label{fig:perp_vs_training_frac}
  \end{subfigure}
  ~
  \begin{subfigure}[t]{1.75in}
    \centering
    \vskip 0pt
    \includegraphics[width=1.75in]{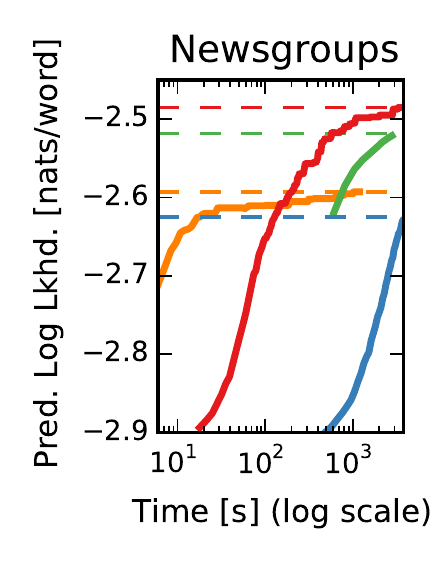}
    \label{fig:perp_vs_time}
  \end{subfigure}
  \\
  \vspace{-3.5em}
  \begin{subfigure}[t]{5.25in}
    \centering
    \vskip 0pt
    \includegraphics[width=\textwidth]{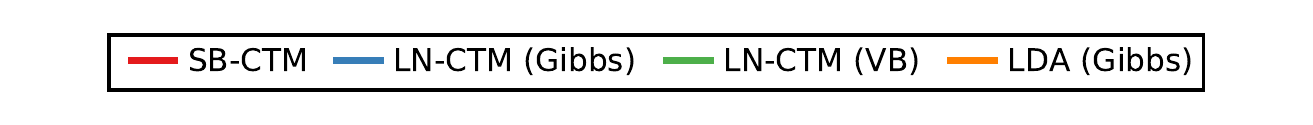}
  \end{subfigure}
  \vspace{-1em}
  %2250 documents
  \caption{A comparison of correlated topic model performance. 
  %The correlated topic model discovers co-occurring topics and leverages these relationships to make improved predictions on new and partial documents. 
  The left panel shows a subset of the inferred topic correlations for the AP News corpus. Two examples are highlighted: a) positive correlation between topics (\textit{house, committee, congress, law}) and 
(\textit{Bush, Dukakis, president, campaign}), and b) anticorrelation between (\textit{percent, year, billion, rate}) and (\textit{court, case, attorney, judge}). The middle and right panels demonstrate the efficacy of our SB-CTM relative to competing models on the AP News corpus and the 20 Newsgroup corpus, respectively. 
  }
\label{fig:ctm}
\end{figure}

The Latent Dirichlet Allocation (LDA) \citep{blei2003latent} is a popular model for learning topics
from text corpora.
The Correlated Topic Model (CTM) \citep{blei2006correlated} extends LDA by including a Gaussian
correlation structure among topics.
This correlation model is powerful not only because it reveals correlations among topics but also because inferring such
correlations can significantly improve predictions, especially when inferring
the remaining words in a document after only a few have been revealed \cite{blei2006correlated}.
However, the addition of this Gaussian correlation structure breaks the
Dirichlet-Multinomial conjugacy of LDA, making estimation and particularly
Bayesian inference and model-averaged predictions more challenging.
An approximate maximum likelihood approach using variational EM \cite{blei2006correlated} is
often effective, but a fully Bayesian approach which integrates out parameters may be preferable, especially when making predictions based on a small number of revealed words in a document.
A recent Bayesian approach based on a P\'{o}lya-Gamma augmentation to the logistic normal
CTM (LN-CTM) \cite{chen2013scalable} provides a Gibbs sampling algorithm with conjugate updates,
but the Gibbs updates are limited to single-site resampling of one scalar at a
time, which can lead to slow mixing in correlated models.

In this section we show that MCMC sampling in a correlated topic model based on the
stick breaking construction (SB-CTM) can be significantly more efficient than
sampling in the LN-CTM while maintaining the same integration advantage over EM.
%Additionally, we provide mean field and SVI algorithms so that such models can
%be scaled to very large datasets.

In the standard LDA model, each topic $\bbeta_t$ ($t=1,2,\ldots,T$) is a
distribution over a vocabulary of $V$ possible words, and each document $d$ has
a distribution over topics $\btheta_d$ ($d=1,2,\ldots,D$).
The $n$th word in document $d$ is denoted $w_{n,d}$ for $d=1,2,\ldots,N_d$.
When each $\bbeta_t$ and $\btheta_d$ is given a symmetric Dirichlet prior with
concentration parameters $\alpha_\beta$ and $\alpha_\theta$, respectively, the
full generative model is
\begin{equation}
  \bbeta_t \sim \distDirichlet(\alpha_\beta), \quad
  \btheta_d \sim \distDirichlet(\alpha_\theta), \quad
  z_{n,d} \given \btheta_d \sim \distCategorical(\btheta_d), \quad
  w_{n,d} \given z_{n,d}, \{\bbeta_t\} \sim \distCategorical(\bbeta_{z_{n,d}}).
\end{equation}
The CTM replaces the Dirichlet prior on each $\btheta_d$ with a new prior that
models the coordinates of~$\btheta_d$ as mutually correlated.
This correlation structure on $\btheta_d$ is induced by first sampling a
correlated Gaussian vector $\bpsi_{d}$ and then applying the logistic normal map:
\begin{equation}
  \bpsi_{d} \given \bmu, \bSigma \sim \mathcal{N}(\bmu, \bSigma), \quad \btheta_d = \piln(\bpsi_{d})
\end{equation}
where the Gaussian parameters $(\bmu, \bSigma)$ can be given a conjugate
normal-inverse Wishart (NIW) prior.
Analogously, our SB-CTM generates the correlation structure by instead applying
the stick-breaking logistic map:
\begin{equation}
  \bpsi_{d} \given \bmu, \bSigma \sim \mathcal{N}(\bmu, \bSigma), \quad \btheta_d = \pisb(\bpsi_{d}).
\end{equation}
The goal is then to infer the posterior distribution over the topics $\bbeta_t$,
the documents' topic allocations~$\bpsi_d$, and their mean and correlation
structure $(\bmu, \bSigma)$.
(In the case of the EM algorithm of \cite{blei2006correlated}, the task is to approximate
maximum likelihood estimates of the same parameters.)
Modeling correlation structure within the topics $\bbeta$ can be done
analogously.

For fully Bayesian MCMC inference in the SB-CTM, we develop a Gibbs sampler
that exploits the block conditional Gaussian structure provided by
the stick-breaking construction.
%We also develop analogous mean field and SVI algorithms in the supplementary
%materials.
The Gibbs sampler iteratively samples
${\bz \given \bw, \bbeta, \bpsi}$; \,
${\bbeta \given \bz, \bw}$; \,
${\bpsi \given \bz, \bmu, \bSigma, \bomega}$; \,
and $\bmu, \bSigma \given \bpsi$; as well as the auxiliary variables~$\bomega \given \bpsi, \bz$.
The first two are standard updates for LDA models, so we focus on the latter
three.
Using the identities derived in Section~\ref{sec:pg-aug}, the conditional density of each $\bpsi_{d} \given \bz_d, \bmu, \bSigma, \bomega$ can be written
\begin{equation}
p(\bpsi_{d} \given \bz_d, \bomega_{d})
\; \propto \;
% \distMultinomial(\bc_d \given N_d, \bpsi_{d}, \omega_{\theta_d}) \distNormal(\psi_{d} \given \mu, \Sigma)
\distNormal(\bpsi \given \kappa(\bc_d), \bOmega_d^{-1})
\;
\distNormal(\bpsi \given \bmu, \bSigma)
\; \propto \;
\distNormal(\bpsi_{d} \given \widetilde{\bmu}, \widetilde{\bSigma}),
\end{equation}
where
\begin{equation}
\widetilde{\bmu} = \widetilde{\bSigma} \left[\kappa(\bc_d) + \bSigma^{-1} \bmu \right],
\quad
\widetilde{\bSigma} = \left[\bOmega_d + \bSigma^{-1} \right]^{-1},
\quad
c_{d,t} = \sum_{n}\bbI[z_{n,d}=t],
\quad
\bOmega_d = \text{diag}(\bomega_d),
\end{equation}
and so it is resampled as a joint Gaussian.
The correlation structure parameters $\bmu$ and $\bSigma$ with a conjugate NIW
prior are sampled from their conditional NIW distribution.
Finally, the auxiliary variables~$\bomega$ are sampled as P\'{o}lya-Gamma random
variables, with $\bomega_{d} \given \bz_d, \bpsi_{d} \sim
\distPolyaGamma(N(\bc_d), \bpsi_{d})$.
A feature of the stick-breaking construction is that the the auxiliary variable
update can be performed in an embarrassingly parallel computation.

We compare the performance of this Gibbs sampling algorithm for the SB-CTM to
the Gibbs sampling algorithm of the LN-CTM \cite{chen2013scalable}, which uses a
different P\'{o}lya-gamma augmentation, as well as the original variational EM
algorithm for the CTM and collapsed Gibbs sampling in standard LDA.
Figure~\ref{fig:ctm} shows results on both the AP News dataset and the 20 Newsgroups
dataset, where models were trained on a random subset of 95\% of the complete
documents and tested on the remaining 5\% by estimating held-out likelihoods of
half the words given the other half.
%(See the supplementary materials for setup details.)
The collapsed Gibbs sampler for LDA is fast but because it does not model
correlations its ability to predict is significantly constrained.
The variational EM algorithm for the CTM is reasonably fast but its point
estimate doesn't quite match the performance from integrating out parameters
via MCMC in this setting.
The LN-CTM Gibbs sampler continues to improve slowly but is limited by its
single-site updates, while the SB-CTM sampler seems to both mix effectively and
execute efficiently due to its block Gaussian updating.

The SB-CTM demonstrates that the stick-breaking construction and corresponding
P\'{o}lya-Gamma augmentation makes inference in correlated topic models both easy
to implement and computationally efficient.
The block conditional Gaussianity also makes inference algorithms
modular and compositional: the construction immediately extends to
dynamic topic models (DTMs) \cite{blei2006dynamic}, in which the latent $\bpsi_{d}$ evolve
according to linear Gaussian dynamics, and inference can be implemented simply by applying
off-the-shelf code for Gaussian linear dynamical systems (see Section~\ref{sec:multinomial_lds}).
Finally, because LDA is so commonly used as a component of other models (e.g.
for images \cite{wang2008spatial}),  easy, effective, modular inference for CTMs and
DTMs is a promising general tool.

To apply correlated topic models to increasingly large datasets, a stochastic
variational inference (SVI) \citep{hoffman2013stochastic} approach is promising.
In Appendix C, we show that the stick-breaking construction enables an
algorithm based on the P\'{o}lya-gamma augmentation that can work with subsets,
or mini-batches, of data in each iteration.
As with the Gibbs sampler, the conditionally conjugate structure makes the
algorithm easy to derive and implement.

% Case study 2: Gaussian Processes with multinomial observations applied to census data

\begin{figure}[t!]
\centering
% First subfigure is 3/7 the pagewidth
  \begin{subfigure}[t]{.49\textwidth}
    \centering
    \vskip 0pt
    \hspace{-.35in}
    \includegraphics[width=3in]{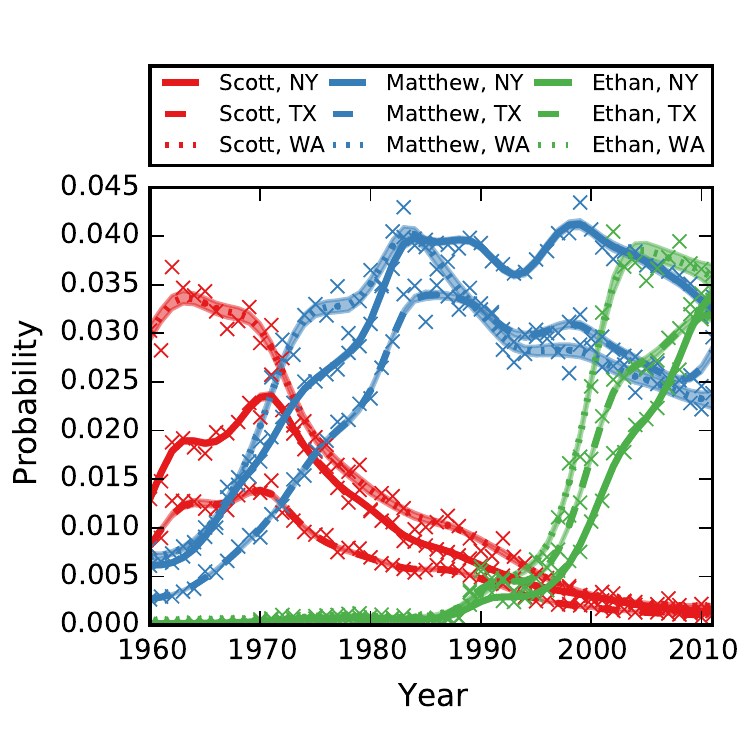}
    \label{fig:census_gp_rate}
  \end{subfigure}
  ~
  \begin{subfigure}[t]{.49\textwidth}
    \centering
    \vskip -3em
    \includegraphics[width=3in]{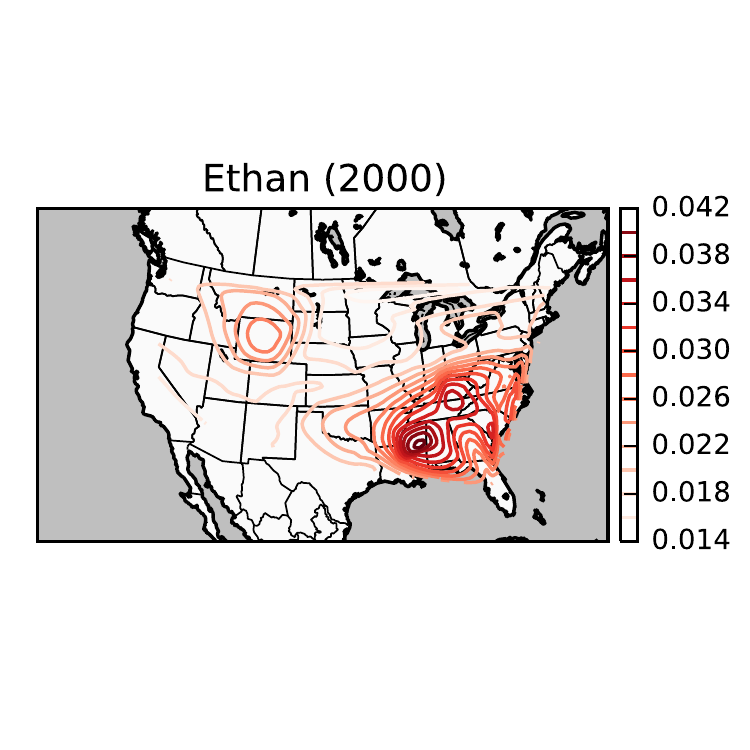}
    \label{fig:ethan_2000}
  \end{subfigure}
  \\
  \vskip-9em
  % Table of results
  \begin{subfigure}[t]{3.2in}
  \centering
  \vspace{-1.25in}
  \begin{tabular}{c|c|c|c|c}
                         & \multicolumn{2}{c|}{\textbf{2012}} & \multicolumn{2}{c}{\textbf{2013}} \\
  \textbf{Model} & \textbf{Top 10} & \textbf{Bot. 10} & \textbf{Top 10} & \textbf{Bot. 10} \\
  \hline
  Static 2011 & 4.2 (1.3) & 0.7 (1.2) & 4.2 (1.4) & 0.8 (1.0) \\
  \hline
  Raw GP & 4.9 (1.1) & 0.7 (0.9) & 5.0 (1.0) & 0.8 (0.9) \\
  \hline
  LNM GP & 6.7 (1.4) & 4.8 (1.7) & 6.8 (1.4) & 4.6 (1.7) \\
  \hline
  SBM GP & 7.3 (1.0) & 4.0 (1.8) & 7.0 (1.0) & 3.9 (1.4) \\
  \end{tabular}
  \vskip.5em
  Average number of names correctly predicted
  \end{subfigure}
  ~
  \begin{subfigure}[t]{2in}
  \centering
    \includegraphics[width=2in]{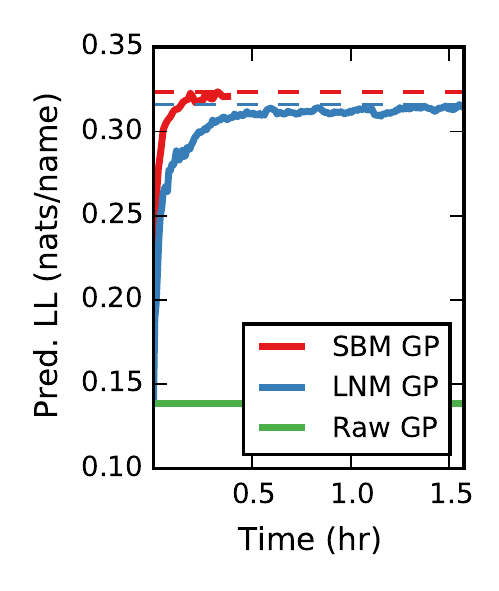}
    \label{fig:census_pred_ll_vs_time}
  \end{subfigure}
  
  \vskip-2em
  \caption{A spatiotemporal Gaussian process applied to the names of children born in the United States from 1960-2013. With a 
limited dataset of only 50 observations per state/year, the stick breaking and logistic normal multinomial GPs (SBM-GP and LNM-GP) outperform na\"ive approaches in 
predicting the top and bottom 10 names (bottom left, parentheses: std. error). Our SBM-GP, which leverages the P\'{o}lya-gamma augmentation, is considerably more 
efficient than the non conjugate LNM-GP (bottom right).  }
\label{fig:census}
\vspace{-1.5em}
\end{figure}

\section{Gaussian processes with multinomial observations}
Consider the United States census data, which lists the first names of children born in each state for the years 1910-2013. Suppose we wish to predict the 
probability of a particular name in New York State in the years 2012 and 2013 given observed names in earlier years. We might reasonably expect that name probabilities 
vary smoothly over time as names rise and fall in popularity, and that name probability would be similar in neighboring states. A Gaussian process naturally captures these 
prior intuitions about spatiotemporal correlations, but the observed name counts are most naturally modeled as multinomial draws from latent probability distributions over 
names for each combination of state and year. We show how efficient inference can be performed in this otherwise difficult model by leveraging the P\'{o}lya-gamma 
augmentation.

%Let~$\bZ \in \reals^{M\times D}$ denote the matrix of~$D$ dimensional inputs and~$\bX \in \naturals^{M \times K}$ denote the observed~$K$ dimensional count vectors 
%for each input. In our example, each row~$\bz_m$ of~$\bZ$ corresponds to the year, latitude, and longitude of an observation, and~$K$ is the number of names. Let~$
%\bPsi(\bZ) \in \reals^{M \times K-1}$ denote the stick-breaking transformation of the name probability vectors for each input point.  The~$k$-th column of this matrix,~$
%\bpsi_k(\bZ)$, is modeled by a Guassian process with mean~$\bmu_k(\bZ)$ and covariance~$K(\bZ, \bZ)$, and the covariance matrix is shared by all the columns. The 
%probability vector for a given input is then,~$\bpi(\bz_m) = \bpi(\left[\psi_1(\bz_m), \ldots, \psi_{K-1}(\bz_m)\right])$, and the joint model is,
%\begin{align*}
%	\bpsi_k(\bZ) &\sim \mathcal{GP}(\bmu_k(\bZ), K(\bZ, \bZ)) & \text{ for } k &\in \{1, \ldots, K-1\} \\
%	\bx(\bz_m) &\sim \distMultinomial(N_m, \pisb(\bz_m)) & \text{ for } m & \in \{1,\ldots, M\}.
%\end{align*}

Let~$\bZ \in \reals^{M\times D}$ denote the matrix of~$D$ dimensional inputs and~$\bX \in \naturals^{M \times K}$ denote the observed~$K$ dimensional count vectors 
for each input. 
In our example, each row~$\bz_m$ of~$\bZ$ corresponds to the year, latitude, and longitude of an observation, and~$K$ is the number of names.
% Let~$\bPsi(\bZ) \in \reals^{M \times K-1}$ denote the stick-breaking transformation of the name probability vectors for each input point.  The~$k$-th column of this matrix,~$\bpsi_k(\bZ)$, is modeled by a Guassian process with mean~$\bmu_k(\bZ)$ and covariance~$K(\bZ, \bZ)$, and the covariance matrix is shared by all the columns. 
Underlying these observations we introduce a set of latent variables,~$\psi_{m,k}$ such that the probability vector at input~$\bz_m$ is~$\bpi_m=\pisb(\bpsi_{m,:})$.
The auxiliary variables for the~$k$-th name,~$\bpsi_{:,k}$, are linked via a Gaussian process with covariance matrix,~$\bC$, whose entry~$C_{i,j}$ is the covariance between input~$\bz_i$ and~$\bz_j$ under the GP prior, and mean vector~$\bmu_k$. The covariance matrix is shared by all names, and the mean is empirically set to match the measured name probability.
% The probability distribution over names for a given input is then,~$\bpi(\bpsi_m) = \bpi(\left[\psi_1(\bz_m), \ldots, \psi_{K-1}(\bz_m)\right])$, and the joint model is,
The full model is then,
\begin{align*}
	\bpsi_{:,k} &\sim \mathcal{GP}(\bmu_k, \bC) & &\text{ for } k \in \{1, \ldots, K-1\} \\
	\bx_{m} &\sim \distMultinomial(N_m, \pisb(\bpsi_{m,:})) & &\text{ for } m \in \{1,\ldots, M\}.
\end{align*}

To perform inference, introduce auxiliary P\'{o}lya-gamma variables,~$\omega_{m,k}$ for each~$\psi_{m,k}$. Conditioned on these variables, the conditional 
distribution of~$\bpsi_{:,k}$ is,
\begin{gather*}
\bpsi_{:,k} \given \bZ, \bX, \bomega, \bmu, \bC \propto 
\distNormal \left(\bpsi_{:,k} \,\bigg|\, \bOmega_{k}^{-1} \kappa(\bX_{:,k}),\, \bOmega_{k}^{-1} \right) \distNormal(\bpsi_{:,k} \given \bmu_k, \bC) \propto 
\distNormal \left(\bpsi_{:,k} \given \widetilde{\bmu}_k, \widetilde{\bSigma}_k \right) \\
\widetilde{\bSigma}_k = \left(\bC^{-1} + \bOmega_k\right)^{-1} \qquad
\widetilde{\bmu}_k = \widetilde{\bSigma}_{k} \left(\bC^{-1}\bmu_k + \kappa(\bX_{:,k})\right),
\end{gather*}
where~$\bOmega_k = \text{diag}(\bomega_{:,k})$. The auxiliary variables are updated according to their conditional distribution:~$\omega_{m,k} \given \bx_m, \psi_{m,k}\sim 
\distPolyaGamma(N_{m,k}, \psi_{m,k})$, where~$N_{m,k}=N_m - \sum_{j<k} x_{m,j}$.

Figure~\ref{fig:census} illustrates the power of this approach on U.S.\ census data. The top two plots show the inferred probabilities under our stick-breaking 
multinomial GP model for the full dataset. Interesting spatiotemporal correlations in name probability are uncovered. In this large-count regime, the posterior uncertainty is negligible 
since we observe thousands of names per state and year, and simply modeling the 
transformed empirical probabilities with a GP works  well. However, in the sparse data regime with only~${N_m=50}$ observations per input, it greatly improves performance to model uncertainty in the latent probabilities using a Gaussian process with multinomial observations.

The bottom panels compare four methods of predicting future names in the years 2012 and 2013 for a down-sampled dataset with~${N_m=50}$: predicting based on the 
empirical probability measured in 2011; a standard GP to the empirical probabilities transformed by~$\pisb^{-1}$ (Raw GP); a GP whose outputs are transformed by the logistic normal 
function,~$\piln$, to obtain multinomial probabilities (LNM GP) fit using elliptical slice sampling \cite{murray-adams-mackay-2010a}; and our stick-breaking multinomial GP (SBM GP). In terms of ability to predict the top and bottom 10 names, the 
multinomial models are both comparable and vastly superior to the naive approaches. 

In terms of efficiency, our model is considerably faster than the logistic normal version, as shown in the bottom right panel. This difference is due to two reasons. First, our 
augmented Gibbs sampler is more efficient than the elliptical slice sampling algorithm used to handle the nonconjugacy in the LNM GP. Second, and perhaps most 
important, we are able to make collapsed predictions in which we compute the predictive distribution test~$\bpsi$'s given $\bomega$, integrating out the training~$\bpsi$. In contrast, the LNM-GP must condition on the training GP values in order to make predictions, and effectively integrate over training samples 
using MCMC. Appendix B goes into greater detail on how marginal predictions are computed and why they are more efficient than predicting conditioned on a single value of~$\bpsi$.

% Case study 3
\section{Multinomial linear dynamical systems}
\label{sec:multinomial_lds}

\begin{figure}
\centering
  % Predictive log likelihood bar charts
  \begin{subfigure}[t]{1.in}
    \centering
    \vskip 0pt
    \includegraphics[width=1.15in]{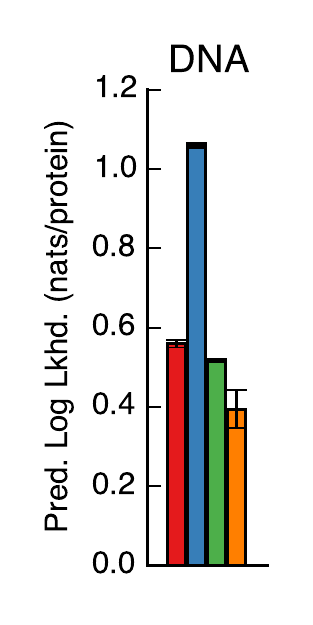}
    \label{fig:dna_lds_1}
  \end{subfigure}
  ~
  \begin{subfigure}[t]{1.in}
    \centering
    \vskip 0pt
    \includegraphics[width=1.15in]{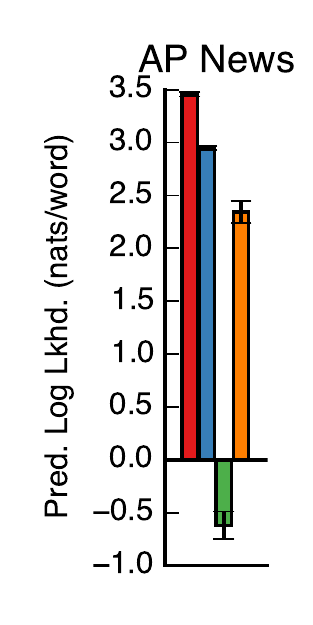}
    \label{fig:ap_predll}
  \end{subfigure}
  ~
  \begin{subfigure}[t]{1.in}
    \centering
    \vskip 0pt
    \includegraphics[width=1.15in]{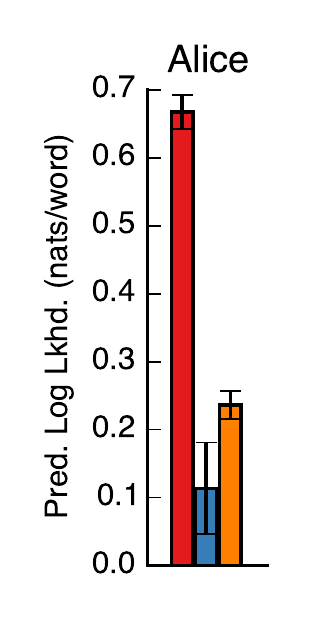}
    \label{fig:dna_lds_1}
  \end{subfigure}
  ~
  \begin{subfigure}[t]{2.in}
    \centering
    \vskip 0pt
    \includegraphics[width=2.10in]{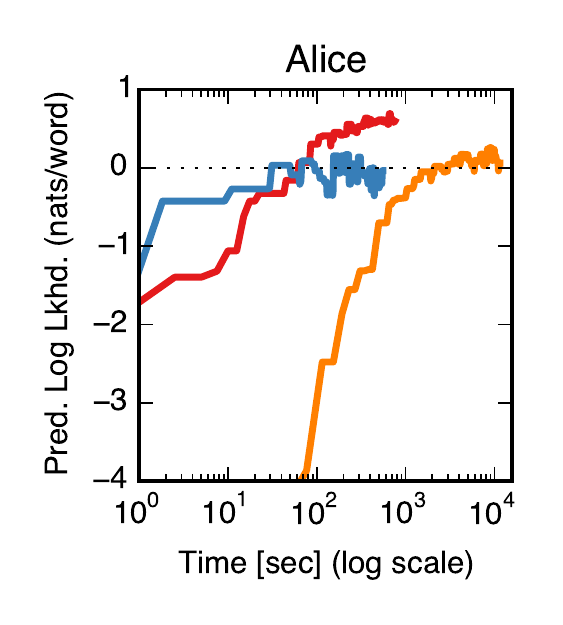}
    \label{fig:dna_lds_1}
  \end{subfigure}
  \\
  \vspace{-2em}
  \begin{subfigure}[t]{5.25in}
    \centering
    \vskip 0pt
    \includegraphics[width=5.25in]{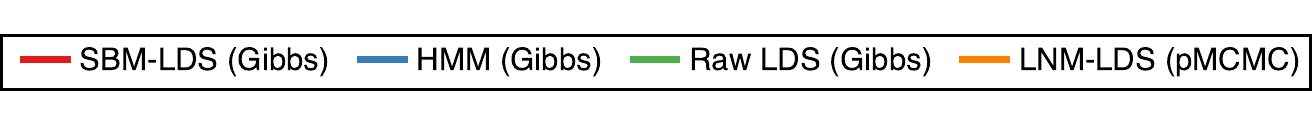}
  \end{subfigure}
  
  \vspace{-1em}
  \caption{Predictive log likelihood comparison of time series models with
    multinomial observations.}
\label{fig:lds}
\vspace{-1.5em}
\end{figure}

While discrete-state hidden Markov models (HMMs) are ubiquitous for modeling
time series and sequence data, it can be preferable to use a continuous state
space model.
In particular, while discrete states have no intrinsic structure, continuous 
states can correspond to a natural embedding or geometry \cite{belanger2015lds}.
These considerations are particularly relevant to text, where word embeddings
\cite{collobert2008unified} have proven to be a powerful tool.

Gaussian linear dynamical systems (LDS) provide very efficient learning and
inference algorithms, but they can typically only be applied when the
observations are themselves linear with Gaussian noise.
While it is possible to apply a Gaussian LDS to count vectors
\cite{belanger2015lds}, the resulting model is misspecified in the sense that,
as a continuous density, the model assigns zero probability to training and
test data.
However, \citet{belanger2015lds} show that this model can still be used for
several machine learning tasks with compelling performance, and that the
efficient algorithms afforded by the misspecified Gaussian assumptions confer a
significant computational advantage.
Indeed, the authors have observed that such a Gaussian model is ``worth
exploring, since multinomial models with softmax link functions prevent
closed-form M step updates and require expensive'' computations
\cite{belanger2014lds}; this paper aims to help bridge precisely this gap and
enable efficient Gaussian LDS computational methods to be applied while
maintaining multinomial emissions and an asymptotically unbiased representation
of the posteiror.
While there are other approximation schemes that effectively extend some of the
benefits of LDSs to nonlinear, non-Gaussian settings, such as the extended
Kalman filter (EKF) and unscented Kalman filter (UKF) \cite{wan2000unscented,
thrun2005probabilistic}, these methods do not allow for asymptotically unbiased
Bayesian inference and can have complex behavior.
Alternatively, particle MCMC (pMCMC) \cite{andrieu2010particle} with ancestor
resampling \cite{lindsten2012ancestor} is a very powerful algorithm that
provides unbiased Bayesian inference for very general state space models, but
it does not enjoy the efficient block updates or conjugacy of LDSs or HMMs.

In this section we show how to use the stick-breaking construction and its
P\'{o}lya-gamma augmentation to perform efficient inference in LDS with multinomial
emissions.
We focus on a Gibbs sampler with fully conjugate updates that utilizes standard
LDS message passing for efficient block updates.

The stick-breaking multinomial linear dynamical system (SBM-LDS) generates
states according to a standard linear Gaussian dynamical system but generates
multinomial observations using the stick-breaking logistic map:
\begin{equation*}
	\bz_0 | \bmu_0, \bSigma_0 \sim \distNormal(\bmu_0, \bSigma_0), \quad
	\bz_t | \bz_{t-1}, \bA, \bB \sim \distNormal(\bA \bz_{t-1}, \bB), \quad
	\bx_t | \bz_t, \bC \sim \distMultinomial(N_t, \pisb(\bC \bz_t)),
\end{equation*}
where $\bz_t \in \reals^D$ is the system state at time $t$ and $\bx_t \in
\naturals^K$ are the multinomial observations.
We suppress notation for conditioning on $\bA$, $\bB$, $\bC$, $\bmu_0$, and
$\bSigma_0$, which are system parameters of appropriate sizes that are given
conjugate priors.
The logistic normal multinomial LDS (LNM-LDS) is defined analogously but uses
$\piln$ in place of $\pisb$.

To produce a Gibbs sampler with fully conjugate updates, we augment the
observations with P\'{o}lya-gamma random variables $\omega_{t,k}$.
As a result, the conditional state sequence $\bz_{1:T} | \bomega_{1:T}, \bx_{1:T}$
is jointly distributed according to a Gaussian LDS in which the diagonal
observation potential at time~$t$ is~$\distNormal(\bOmega_t^{-1} \kappa(\bx_t) | \bC
\bz_t, \bOmega_t^{-1})$;
because the observation potential is diagonal, this block update can be
performed in only $\mathcal{O}(T D^3 + T D^2 K)$ time, and so
these updates can be scaled efficiently to large observation dimension $K$.
Thus the state sequence can be jointly sampled using
off-the-shelf LDS software, and the system parameters can similarly be updated
using standard algorithms.
The only remaining update is to the auxiliary variables, which are sampled
according to ${\bomega_{t} | \bz_t, \bC, \bx \sim \distPolyaGamma(N(\bx_t), \bC
\bz_t)}$.

We compare the SBM-LDS and the Gibbs sampling inference algorithm to three
baseline methods: an LNM-LDS using pMCMC and ancestor resampling for inference,
an HMM using Gibbs sampling, and a ``raw'' LDS which treats the multinomial
observation vectors as observations in $\reals^K$, as in \citet{belanger2015lds}.
We examine each method's performance on each of three experiments: in modeling
a sequence of 682 amino acids from human DNA with 22 dimensional observations, a
set of 20 random AP news articles with an average of 77 words per article and a
vocabulary size of 200 words, and an excerpt of 4000 words from Lewis Carroll's
\emph{Alice's Adventures in Wonderland} with a vocabulary of 1000 words. We
reserved the final 10 amino acids, 10 words per news article, and 100 words
from \emph{Alice} for computing predictive likelihoods.
Each linear dynamical model had a 10-dimensional state space, while the HMM had
10 discrete states (HMMs with 20, 30, and 40 states all performed worse on
these tasks).

Figure~\ref{fig:lds} (left panels) shows the predictive log likelihood for each
method on each experiment, normalized by the number of counts in the test
dataset and setting to zero the likelihood under a multinomial model fit to the
training data mean.
For the DNA data, which has the smallest ``vocabulary'' size, the HMM achieves
the highest predictive likelihood, but the SBM-LDS edges out the other LDS
methods.
On the two text datasets, the SBM-LDS outperforms the other methods,
particularly in \emph{Alice} where the vocabulary is larger and the document is
longer.
In terms of run time, the SBM-LDS is orders of magnitude faster than the
LNM-LDS with pMCMC (right panel) because it mixes much more efficiently over
the latent trajectories.

The SBM-LDS is an easy but powerful linear state space model for multinomial
observations.
The Gibbs sampler leveraging the P\'{o}lya-gamma augmentation appears very
efficient, performing comparably to an optimized HMM implementation and orders
of magnitude faster than a general pMCMC algorithm.
Because the augmentation renders the states' conditional distribution a Gaussian
LDS, it easily interfaces with high-performance LDS software, and extending
these models with additional structure or covariates can be similarly modular.
\section{Related Work}
The stick-breaking transformation used herein was applied to categorical models by \citet{khan2012stick}, 
but they used local variational bound instead of the P\'{o}lya-gamma augmentation. Their promising results 
corroborate our findings of improved performance using this transformation. Their generalized expectation-maximization 
algorithm is not fully Bayesian, and does not integrate into existing Gaussian modeling code as easily as our augmentation.

Conversely, \citet{chen2013scalable} used the P\'{o}lya-gamma augmentation in conjunction with the logistic normal transformation 
for correlated topic modeling, exploiting the conditional conjugacy of a single entry~${\psi_k \given \omega_k, \bpsi_{\neg k}}$ with a 
Gaussian prior. Unlike our stick-breaking transformation which admits block Gibbs sampling over the entire vector~$\bpsi$ simultaneously, 
their approach is limited to single-site Gibbs sampling. As shown in our correlated topic model experiments, this has dramatic effects 
on inferential performance. Moreover, it precludes analytical marginalization and integration with existing Gaussian modeling algorithms. 
For example, it is not immediately applicable to inference in linear dynamical systems with multinomial observations.

\section{Conclusion}
These case studies demonstrate that the stick-breaking multinomial model construction paired with the P\'{o}lya-gamma augmentation yields a flexible class of models with easy, efficient, and compositional inference.
In addition to making these models easy, the methods developed here can also enable new models for multinomial and mixed data: the latent continuous structures used here to model correlations and state-space structure can be leveraged to explore new models for interpretable feature embeddings, interacting time series, and dependence with other covariates.

\paragraph{Acknowledgments}
We thank the members of the Harvard Intelligent Probabilistic Systems (HIPS) group, especially Yakir Reshef, for many helpful conversations. S.W.L. is supported by the Center for Brains, Minds and Machines (CBMM), funded by NSF STC award CCF-1231216. M.J.J. is supported by the Harvard/MIT Joint Research Grants Program. R.P.A. is partially supported by NSF IIS-1421780.

\bibliography{arxiv}

\begin{thebibliography}{21}
\providecommand{\natexlab}[1]{#1}
\providecommand{\url}[1]{\texttt{#1}}
\expandafter\ifx\csname urlstyle\endcsname\relax
  \providecommand{\doi}[1]{doi: #1}\else
  \providecommand{\doi}{doi: \begingroup \urlstyle{rm}\Url}\fi

\bibitem[Andrieu et~al.(2010)Andrieu, Doucet, and
  Holenstein]{andrieu2010particle}
Christophe Andrieu, Arnaud Doucet, and Roman Holenstein.
\newblock Particle {M}arkov chain {M}onte {C}arlo methods.
\newblock \emph{Journal of the Royal Statistical Society: Series B (Statistical
  Methodology)}, 72\penalty0 (3):\penalty0 269--342, 2010.

\bibitem[Murray et~al.(2010)Murray, Adams, and
  MacKay]{murray-adams-mackay-2010a}
Iain Murray, Ryan~P. Adams, and David~J.C. MacKay.
\newblock Elliptical slice sampling.
\newblock \emph{Journal of Machine Learning Research: Workshop and Conference
  Proceedings (AISTATS)}, 9:\penalty0 541--548, 05/2010 2010.
\newblock URL
  \url{http://hips.seas.harvard.edu/files/w/papers/murray-adams-mackay-2010a.pdf}.

\bibitem[Polson et~al.(2013)Polson, Scott, and Windle]{polson2013bayesian}
Nicholas~G Polson, James~G Scott, and Jesse Windle.
\newblock Bayesian inference for logistic models using {P}{\'o}lya--gamma
  latent variables.
\newblock \emph{Journal of the American Statistical Association}, 108\penalty0
  (504):\penalty0 1339--1349, 2013.

\bibitem[Zhou et~al.(2012)Zhou, Li, Dunson, and Carin]{zhou2012lognormal}
Mingyuan Zhou, Lingbo Li, David Dunson, and Lawrence Carin.
\newblock Lognormal and gamma mixed negative binomial regression.
\newblock In \emph{Proceedings of the International Conference on Machine
  Learning}, volume 2012, page 1343, 2012.

\bibitem[Ren et~al.(2011)Ren, Du, Carin, and Dunson]{ren2011logistic}
Lu~Ren, Lan Du, Lawrence Carin, and David Dunson.
\newblock Logistic stick-breaking process.
\newblock \emph{The Journal of Machine Learning Research}, 12:\penalty0
  203--239, 2011.

\bibitem[Khan et~al.(2012)Khan, Mohamed, Marlin, and Murphy]{khan2012stick}
Mohammad~E Khan, Shakir Mohamed, Benjamin~M Marlin, and Kevin~P Murphy.
\newblock A stick-breaking likelihood for categorical data analysis with latent
  {G}aussian models.
\newblock In \emph{International Conference on Artificial Intelligence and
  Statistics}, pages 610--618, 2012.

\bibitem[Albert and Chib(1993)]{albert1993bayesian}
James~H Albert and Siddhartha Chib.
\newblock Bayesian analysis of binary and polychotomous response data.
\newblock \emph{Journal of the American statistical Association}, 88\penalty0
  (422):\penalty0 669--679, 1993.

\bibitem[Holmes et~al.(2006)Holmes, Held, et~al.]{holmes2006bayesian}
Chris~C Holmes, Leonhard Held, et~al.
\newblock Bayesian auxiliary variable models for binary and multinomial
  regression.
\newblock \emph{Bayesian Analysis}, 1\penalty0 (1):\penalty0 145--168, 2006.

\bibitem[Blei and Lafferty(2006{\natexlab{a}})]{blei2006correlated}
David Blei and John Lafferty.
\newblock Correlated topic models.
\newblock \emph{Advances in Neural Information Processing Systems},
  18:\penalty0 147, 2006{\natexlab{a}}.

\bibitem[Blei et~al.(2003)Blei, Ng, and Jordan]{blei2003latent}
David~M Blei, Andrew~Y Ng, and Michael~I Jordan.
\newblock Latent {D}irichlet allocation.
\newblock \emph{the Journal of machine Learning research}, 3:\penalty0
  993--1022, 2003.

\bibitem[Chen et~al.(2013)Chen, Zhu, Wang, Zheng, and Zhang]{chen2013scalable}
Jianfei Chen, Jun Zhu, Zi~Wang, Xun Zheng, and Bo~Zhang.
\newblock Scalable inference for logistic-normal topic models.
\newblock In \emph{Advances in Neural Information Processing Systems}, pages
  2445--2453, 2013.

\bibitem[Blei and Lafferty(2006{\natexlab{b}})]{blei2006dynamic}
David~M Blei and John~D Lafferty.
\newblock Dynamic topic models.
\newblock In \emph{Proceedings of the International Conference on Machine
  Learning}, pages 113--120. ACM, 2006{\natexlab{b}}.

\bibitem[Wang and Grimson(2008)]{wang2008spatial}
Xiaogang Wang and Eric Grimson.
\newblock Spatial latent {D}irichlet allocation.
\newblock In \emph{Advances in Neural Information Processing Systems}, pages
  1577--1584, 2008.

\bibitem[Hoffman et~al.(2013)Hoffman, Blei, Wang, and
  Paisley]{hoffman2013stochastic}
Matthew~D Hoffman, David~M Blei, Chong Wang, and John Paisley.
\newblock Stochastic variational inference.
\newblock \emph{The Journal of Machine Learning Research}, 14\penalty0
  (1):\penalty0 1303--1347, 2013.

\bibitem[Belanger and Kakade(2015)]{belanger2015lds}
David Belanger and Sham Kakade.
\newblock A linear dynamical system model for text.
\newblock In \emph{Proceedings of the International Conference on Machine
  Learning}, 2015.

\bibitem[Collobert and Weston(2008)]{collobert2008unified}
Ronan Collobert and Jason Weston.
\newblock A unified architecture for natural language processing: Deep neural
  networks with multitask learning.
\newblock In \emph{Proceedings of the International Conference on Machine
  Learning}, pages 160--167. ACM, 2008.

\bibitem[Belanger and Kakade(2014)]{belanger2014lds}
David Belanger and Sham Kakade.
\newblock Embedding word tokens using a linear dynamical system.
\newblock In \emph{NIPS 2014 Modern ML+NLP Workshop}, 2014.

\bibitem[Wan and Van Der~Merwe(2000)]{wan2000unscented}
Eric~A Wan and Rudolph Van Der~Merwe.
\newblock The unscented {K}alman filter for nonlinear estimation.
\newblock In \emph{Adaptive Systems for Signal Processing, Communications, and
  Control Symposium 2000. AS-SPCC. The IEEE 2000}, pages 153--158. IEEE, 2000.

\bibitem[Thrun et~al.(2005)Thrun, Burgard, and Fox]{thrun2005probabilistic}
Sebastian Thrun, Wolfram Burgard, and Dieter Fox.
\newblock \emph{Probabilistic robotics}.
\newblock MIT press, 2005.

\bibitem[Lindsten et~al.(2012)Lindsten, Sch{\"o}n, and
  Jordan]{lindsten2012ancestor}
Fredrik Lindsten, Thomas Sch{\"o}n, and Michael~I Jordan.
\newblock Ancestor sampling for particle {G}ibbs.
\newblock In \emph{Advances in Neural Information Processing Systems}, pages
  2591--2599, 2012.

\bibitem[Bishop(2006)]{bishop2006prml}
Christopher~M Bishop.
\newblock \emph{Pattern recognition and machine learning}.
\newblock Springer, 2006.

\end{thebibliography}
\bibliographystyle{unsrtnat}

\appendix

\section{Transforming between $p(\bpsi)$ and~$p(\bpi)$}

Since the mapping between~$\bpi$ and~$\bpsi$ is invertible, we can compute the distribution on~$\bpi$ that is implied by a Gaussian distribution on~$\bpsi$. 
Assume~$\bpsi \sim \distNormal(\bmu, \bSigma)$. Then,
\begin{align*}
  p(\bpi \given \bmu, \bSigma) &= \distNormal(\pisb^{-1}(\bpi) \given \bmu, \bSigma) \left| \frac{\mathrm{d} \bpsi}{\mathrm{d} \bpi} \right| \\
%  &= \distNormal(h^{-1}(\bpi) \given \bmu, \bSigma) \left| \begin{bmatrix}
%    \frac{\partial \psi_1}{\partial \pi_1} & \ldots & \frac{\partial \psi_1}{\partial \pi_{K-1}} \\
%    & & \\
%    \frac{\partial \psi_{K-1}}{\partial \pi_1} & \ldots & \frac{\partial \psi_{K-1}}{\partial \pi_{K-1}} 
%    \end{bmatrix}
% \right| \\
% &= \distNormal(h^{-1}(\bpi) \given \bmu, \bSigma) \cdot |\bJ|
\end{align*}

From above, we have
\begin{align*}
  \psi_1 &= \sigma^{-1}(\pi_1), & 
  \psi_2 &= \sigma^{-1}\left(\frac{\pi_2}{1-\pi_1}\right), &
  & \ldots, & 
  \psi_{k} &= \sigma^{-1} \left( \frac{\pi_k}{1-\sum_{j<k} \pi_j} \right).
\end{align*}
Let~
\begin{align*}
  g(x)=\frac{\mathrm{d} \sigma^{-1}(\mathit{x})}{\mathrm{d} \mathit{x}} \bigg|_{\mathit{x}=x}= \frac{\mathrm{d}}{\mathrm{d} x} \log \left(\frac{x}{1-x} \right) = \frac{1}{x} + 
\frac{1}{1-x} = \frac{1}{x(1-x)}.
\end{align*}
Then,
\begin{align*}
  \frac{\partial \psi_1}{\partial \pi_1} &= g(\pi_1), &
  \frac{\partial \psi_k}{\partial \pi_{k}} &= g\left(\frac{\pi_k}{1-\sum_{j<k}\pi_j}\right) \frac{1}{1-\sum_{j<k} \pi_j}, &
  \frac{\partial \psi_k}{\partial \pi_{j>k}} &= 0.
\end{align*}
Since the Jacobian of the inverse transformation is lower diagonal, its determinant is simply the product of its diagonal,
\begin{align*}
  \left| \frac{\mathrm{d} \bpsi}{\mathrm{d}\bpi} \right| &= \prod_{k=1}^K \left[g\left(\frac{\pi_k}{1-\sum_{j<k}\pi_j}\right) \frac{1}{1-\sum_{j<k} \pi_j} \right] \\
          &= \prod_{k=1}^K \left[\frac{1 - \sum_{j < k} \pi_j}{\pi_k} \frac{1-\sum_{j<k}\pi_j}{1-\sum_{j<k}\pi_j -\pi_k} \frac{1}{1-\sum_{j<k} \pi_j} \right] \\
          &= \prod_{k=1}^K \left[\frac{1 - \sum_{j=1}^{k-1} \pi_j}{\pi_k (1-\sum_{j=1}^{k} \pi_j)} \right]
\end{align*}
Thus, the final density is,
\begin{align*}
p(\bpi \given \bmu, \bSigma) &= \distNormal(\pisb^{-1}(\bpi) \given \bmu, \bSigma) \cdot \prod_{k=1}^K \left[\frac{1 - \sum_{j=1}^{k-1} \pi_j}{\pi_k (1-\sum_{j=1}^{k} \pi_j)} \right].
\end{align*}

% p(psi | \alpha): Dirichlet to distribution over psi
Now, suppose we are given a Dirichlet distribution,~$\bpi\sim \distDirichlet(\bpi \given \balpha)$, and we wish to compute the density on~$\bpsi$. We have,
\begin{align*}
p(\bpsi \given \balpha) &= \distDirichlet(\pisb(\bpsi) \given \balpha) \cdot \left| \frac{\mathrm{d} \bpi}{\mathrm{d}\bpsi}\right|\\
&= \distDirichlet(\pisb(\bpsi) \given \balpha) \cdot \prod_{k=1}^K \left[\frac{\pi_k (1-\sum_{j=1}^{k} \pi_j)}{1 - \sum_{j=1}^{k-1} \pi_j} \right],
\end{align*}
where we have used the fact that the Jacobian of the inverse transformation is simply the inverse of the Jacobian of the forward transformation.
 We simply need to rewrite the Jacobian in terms of~$\psi$ rather than~$\pi$. 
Note that~$1-\sum_{j<k} \pi_j$ is the length of the remaining stick and~$\sigma(\psi_k)$ is the fraction of the remaining ``stick'' allocated to~$\pi_k$.  Thus, the remaining 
stick length is equal to,
\begin{align*}
1-\sum_{j<k} \pi_j &\equiv \prod_{j<k} (1-\sigma(\psi_j)) \equiv \prod_{j<k} \sigma(-\psi_j).
\end{align*}
Moreover,~$\pi_k=\sigma(\psi_k) (1-\sum_{j<k} \pi_j)=\sigma(\psi_k)\prod_{j<k} \sigma(-\psi_j)$. Thus,
\begin{align*}
p(\bpsi \given \balpha) &= \distDirichlet(\pisb(\bpsi) \given \balpha) \cdot \prod_{k=1}^K \left[\frac{\left( \sigma(\psi_k) \prod_{j< k} \sigma(-\psi_j)\right)  \left(\prod_{j \leq k} \sigma(-\psi_j)\right) }{\prod_{j<k} \sigma(-\psi_j)} \right],\\
 &= \distDirichlet(\pisb(\bpsi) \given \balpha) \cdot \prod_{k=1}^K \left[\sigma(\psi_k) \prod_{j \leq k} \sigma(-\psi_j) \right],
\end{align*}
Expanding the Dirichlet distribution and substituting~$\psi$ for~$\pi$, we conclude that,
\begin{align*}
p(\bpsi \given \balpha) &= \frac{1}{\mathrm{B}(\balpha)} \prod_{k=1}^{K-1} \sigma(\psi_k)^{\alpha_k} \cdot \sigma(-\psi_k)^{\sum_{j=k+1}^{K} \alpha_j}.
\end{align*}
This factorized form is unsurprising given that the Dirichlet distribution can be written as a stick-breaking product of beta distributions in the same way that the multinomial can be written as a product of binomials. Each term in the product above corresponds to the transformed beta distribution over~$\widetilde{\pi}_k$. 

\begin{figure}
\centering
  \begin{subfigure}[t]{5in}
    \centering
    \vskip 0pt
    \includegraphics[width=\textwidth]{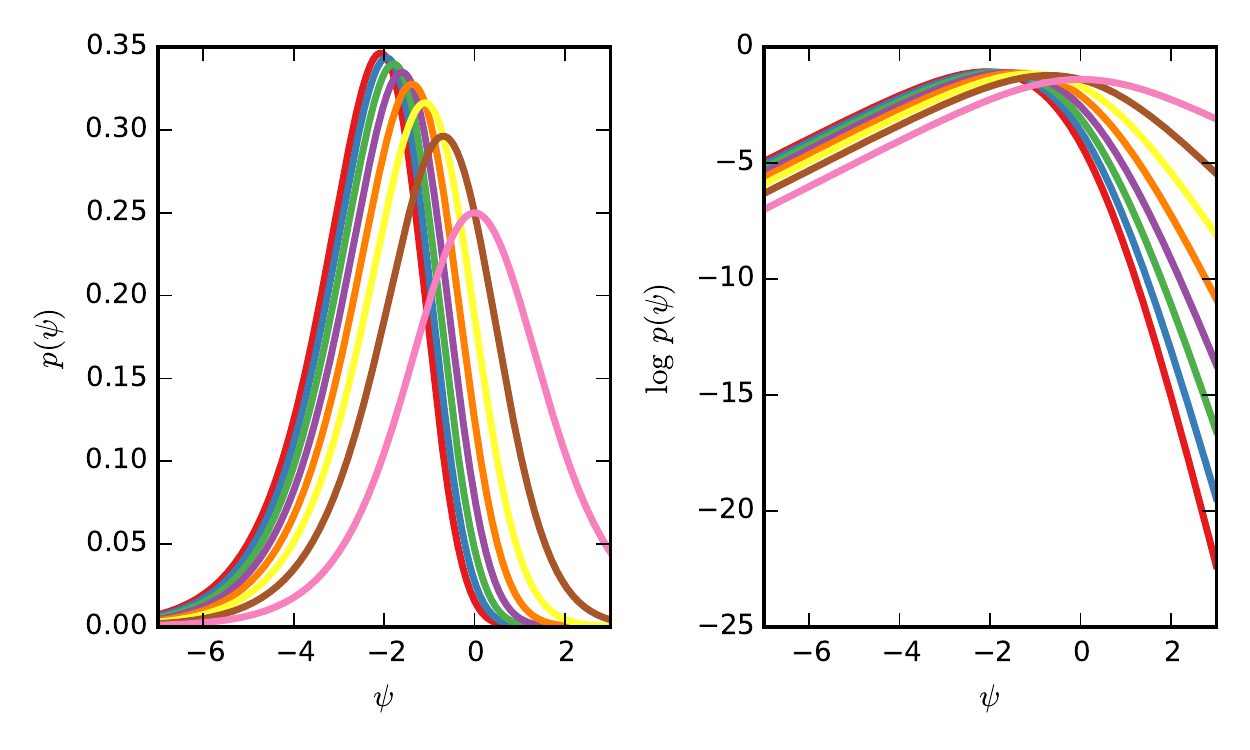}
    \label{fig:1b_psi}
  \end{subfigure}
  \\
  \begin{subfigure}[t]{5in}
    \centering
    \vskip -2em
    \includegraphics[width=\textwidth]{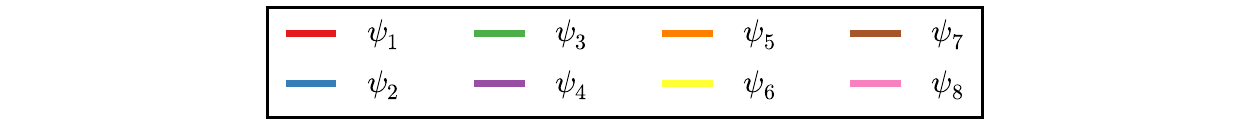}
  \end{subfigure}
  \caption{Density and log density of~$p(\bpsi \given \balpha=\bone)$, the density on~$\bpsi$ implied by a~$K=9$ dimensional symmetric Dirichlet density on~$\bpi$ with parameter~$\alpha=1$.}
  \label{fig:psi_density}
\end{figure}

Figure~\ref{fig:psi_density} shows the marginal densities on~$\psi_k$ implied by a~$K=9$ dimensional symmetric Dirichlet prior on~$\bpi$ with~$\alpha=1$. The densities of~$\psi_k$ become increasingly skewed for small values of~$k$, but they are still well approximate by a Gaussian distribution. In order to approximate a uniform distribution, we numerically compute the mean and variance of these densities to set the parameters of a diagonal Guassian distribution.

\section{Marginal Predictions with the Augmented Model}
\begin{figure}
\centering
  \begin{subfigure}[t]{3in}
    \centering
    \vskip 0pt
    \includegraphics[width=\textwidth]{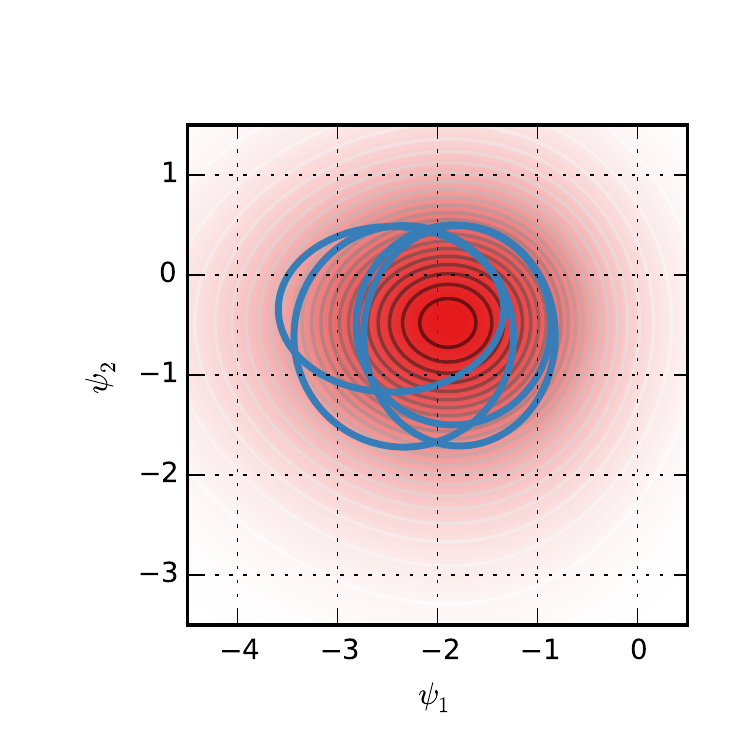}
    \label{fig:1b_psi}
  \end{subfigure}
  \vskip-2em
  \caption{Marginal density,~$p(\bpsi \given \bx)$ in red shading along with the ellipses of multivariate normal conditional distribution~$p(\bpsi \given \bx, \bomega)$ for 4 steps of the Gibbs sampler. 
In Gaussian models where we aim to predict~$\bpsi_{\mathsf{test}}$ on test data, there are substantial gains to be had from making marginal predictions of~$\bpsi_{\mathsf{test}} \given \bx, \bomega$, integrating out~$\bpsi_{\mathsf{train}}$. 
The key is that the conditional densities overlap substantially with the marginal density.
} 
  \label{fig:psi_density}
\end{figure}

One of the primary advantages offered by the P\'{o}lya-gamma augmentation is the ability to make marginal predictions about~$\bpsi_{\mathsf{test}} \given \bx, \bomega$, integrating out the value of~$\bpsi_{\mathsf{train}}$. For example, in the GP multinomial regression models described in the main text, the methods were evaluated on the accuracy of their predictions about future name probabilities, which were functions of~$\bpsi_{\mathsf{test}}$. When~$p(\bpsi_{\mathsf{train}})$ and~$p(\bpsi_{\mathsf{test}} \given \bpsi_{\mathsf{train}})$ are both Gaussian, we can integrate out the latent training variables in order to predict their test values. In a latent Gaussian-multinomial model, the posterior distribution over those latent training variables is non-Gaussian, but after P\'{o}lya-gamma augmentation, it is rendered Gaussian. 

With the augmentation, we can write
\begin{align*}
p(\bpsi_{\mathsf{test}} \given \bx)
%&=\frac{1}{M} \sum_{m=1}^M \int p(\bpsi_{\mathsf{test}}, \bpsi_{\mathsf{train}}, \bomega^{(m)} \given \bx)\, \mathrm{d} \bpsi_{\mathsf{train}} \\
&\approx \frac{1}{M} \sum_{m=1}^M \int p(\bpsi_{\mathsf{test}} \given \bpsi_{\mathsf{train}}) \, p(\bpsi_{\mathsf{train}} \given \bx, \bomega^{(m)})\, \mathrm{d} \bpsi_{\mathsf{train}} & \bomega^{(m)} &\sim p(\bomega \given \bx), \\
\end{align*}
and perform Monte Carlo integration over~$\bomega$ in order to compute the predictive distribution. By contrast, in the standard formulation we must perform Monte Carlo integration over~$\bpsi$,
\begin{align*}
p(\bpsi_{\mathsf{test}} \given \bx)
&= \frac{1}{M} \sum_{m=1}^M p(\bpsi_{\mathsf{test}} \given \bpsi_{\mathsf{train}}^{(m)}) & \bpsi_{\mathsf{train}}^{(m)} &\sim p(\bpsi_{\mathsf{train}} \given \bx). \\
\end{align*}
Why does the augmented model confer a predictive advantage? It does not come from performing Monte Carlo integration over a smaller dimension since~$\bomega$ and~$\bpsi_{\mathsf{train}}$ are of the same size. Instead, it comes from the ability of the conjugate Gibbs sampler to efficiently mix over~$\bpsi$ and~$\bomega$, and from the ability of a single sample of~$\bomega$ to render a conditional Gaussian distribution over~$\bpsi$ that captures much of the volume of the true marginal distribution.

This latter point is illustrated in Figure~\ref{fig:psi_density}. The red shading shows the true marginal density of~$\bpsi$ and the blue ellipses show the conditional density for a fixed value of~$\bomega$. Each ellipse capture a significant amount of the marginal distribution, indicating that with a single sample of~$\bomega$ we can integrate over a substantial amount of the uncertainty in~$\bpsi$. This example is only for a~$K=3$ dimensional multinomial observation, but this intuition should extend to higher dimensions in which the advantages of analytical integration should be more readily apparent.

\section{Variational Inference for Correlated Topic Models}
We use the following factorized approximation to the posterior distribution,
\begin{multline*}
p(\{\bpsi_d, \bomega_d\}, \{\bbeta_{t}\}, \{\{z_{n,d} \} \}, \bmu_{\theta}, \bSigma_{\theta}, \bmu_\beta, \bSigma_\beta \given \{\{w_{n,d}\}\}) \\
\approx \left[ \prod_{d=1}^{D} q(\bpsi_d) \prod_{t=1}^T q(\omega_{d,t}) \prod_{n=1}^{N_d} q(z_{n,d}) \right] \left[ \prod_{t=1}^T q(\bbeta_t) \right] q(\bmu_{\theta}, \bSigma_\theta).
\end{multline*}
First let's consider the variational distribution for~$\bpsi_d$ and~$\bomega_d$. 
From the conjugacy of the model, we have
\begin{align*}
q(\bpsi_d) &= \distNormal(\bpsi_d \given \widetilde{\bmu}_{\btheta_d}, \widetilde{\bSigma}_{\btheta_d}) \\
\widetilde{\bmu}_{\btheta_d} &= \widetilde{\bSigma}_{\btheta_d} \left[ \bbE [\bkappa( \bc_d, N_d)] + \bbE[\bSigma_{\theta}^{-1} \bmu_{\theta} ] \right ]^{-1} \\
\widetilde{\bSigma}_{\btheta_d} &= \left[ \bbE [\text{diag}(\bomega_d)] + \bbE[\bSigma_{\btheta}^{-1}] \right ]^{-1}, \\
\end{align*}
and
\begin{align*}
\bbE[(\bkappa(\bc_d, N_d))_t] &= \bbE[c_{d,t} - N_{d,t}/2]  \\
&= \bbE[c_{d,t} - (N_d-\sum_{t' < t}c_{d,t'})/2]  \\
& = \bbE[c_{d,t}] + \frac{1}{2}\sum_{t' < t} \bbE[c_{d,t'}] - \frac{N_d}{2}\\
\bbE[c_{d,t}] &= \sum_{n=1}^{N_d} \bbE[z_{n,d}=t].
\end{align*}

The factor for~$\bomega_d$ is not available in closed form. We have,
\begin{align*}
\log q(\bomega_d) &= \bbE_{\bpsi_d, \bz_{d}}[\log p(\bz_{d} \given \bpsi_d, \bomega_d)] + \text{const.}\\
&= \bbE_{\bpsi_d, \bz_{d}}[\log \distPolyaGamma(\bomega_d \given \bN(\bc_d), \bpsi_{d})] + \text{const.}
\end{align*}
Instead, following \cite{zhou2012lognormal}, we restrict the variational factor over~$\omega$ to take the form of a Polya-gamma distribution,
~${\omega_{d,t}\sim \distPolyaGamma(\omega_{d,t} \given N_{d,t}, \psi_{d,t})}$, where~${N_{d,t}=[N(\bc_d)]_t}$.
To perform the updates for~$\bpsi_d$, we only need the expectations of~$\omega_{d,t}$ under the P\'{o}lya-gamma factors. The mean of~$PG(b,c)$ distribution is available in closed form:~$\bbE_{\omega \sim \distPolyaGamma(b,c)}[\omega]=\frac{b}{2c}\tanh(\frac{c}{2})$. Since the parameters of the P\'{o}lya-gamma distribution have variational factors, we use iterated expectations and Monte Carlo methods to approximate the expectation,
\begin{align*}
\bbE[q(\omega_{d,t})] &= \bbE_{\bpsi_{d,t}, \bz_{d}} \left[ \bbE_{\bomega_{d,t} | \bpsi_{d,t}, \bz_{d}}[ \distPolyaGamma( \omega_{d,t} \given \bN(\bc_d)_t, \bpsi_{d,t})] \right] \\
&= \frac{1}{2}\bbE_{\bz_{d}} \left[ \bN(\bc_d)_t\right]  \bbE_{\psi_{d,t}} \left[\frac{\tanh(\psi_{d,t}/2)}{\psi_{d,t}} \right] \\
&= \frac{1}{2}\left(N_d-\sum_{t'=1}^{t} \bbE_{\bz_{d}} \left[ c_{d,t'} \right] \right) \bbE_{\psi_{d,t}} \left[\frac{\tanh(\psi_{d,t} /2)}{\psi_{d,t}} \right],\\
&= \frac{1}{2}\left(N_d-\sum_{t'=1}^{t-1} \sum_{n=1}^{N_d} \bbE_{\bz_d}[z_{n,d}=t'] \right) \bbE_{\psi_{d,t}} \left[\frac{\tanh(\psi_{d,t} /2)}{\psi_{d,t}} \right].
\end{align*}

The updates for the global topic distribution parameters,~$\bmu_\theta$ and~$\bSigma_\theta$, depend only on their normal inverse-Wishart prior and the expectations with respect to~$q(\bpsi_{d})$. These follow their standard form, see, for example, \citet{bishop2006prml}.

The variational updates for~$z_{n,d}$ and~$\bbeta_t$ are straightforward. 
\begin{align*}
\log q(z_{n,d}) &= \bbE_{\btheta_d, \bbeta} \left[ \log p(w_{n,d} \given z_{n,d}, \btheta_d, \{\bbeta_t\}) \right] + \text{ const.}\\
&= \bbE_{\btheta_d, \bbeta} \left[ \sum_{t=1}^T z_{n,d} (\log \beta_{t,w_{d,n}} + \log \theta_{d,t}) \right] + \text{ const.}\\
&= \sum_{t=1}^T z_{n,d} (\bbE_{\bbeta}[\log \beta_{t,w_{n,d}}] + \bbE_{\btheta_d}[\log \theta_{d,t}]) + \text{ const.}
\end{align*}
This implies that~$q(z_{n,d})$ is categorical with parameters,
\begin{align*}
q(z_{n,d}) &= \distCategorical(z_{n,d} \given \widetilde{\bu}_{n,d}), \\
\widetilde{u}_{n,d,t} &= \frac{1}{Z} \exp\left\{\bbE_{\bbeta}[\log \beta_{t,w_{n,d}}] + \bbE_{\btheta_d}[\log \theta_{d,t}]\right\}, \\
Z &= \sum_{t'=1}^T \exp\left\{\bbE_{\bbeta}[\log \beta_{t',w_{n,d}}] + \bbE_{\btheta_d}[\log \theta_{d,t'}]\right\}
\end{align*}
The challenge is that~$\bbE_{\btheta_d}[\log \theta_{d,t}]$ is not available in closed form. Instead we must approximate it by Monte Carlo sampling the corresponding value of~$\bpsi_d$.

Last,
\begin{align*}
\log q(\bbeta_t) &= \bbE \left[ \log p(\bw \given \bz, \btheta) + \log p(\bbeta_t \given \balpha) \right] + \text{ const.} \\
&= \sum_{d=1}^D \sum_{n=1}^{N_d} \bbE[z_{n,d}=t] \log \beta_{t, w_{n,d}} + \sum_{v=1}^V (\alpha_v -1) \log \beta_{t,v} + \text{ const.}
\end{align*}
We recognize this as a Dirichlet distribution,
\begin{align*}
q(\bbeta_t) &= \distDirichlet(\bbeta_t \given \widetilde{\balpha}_t), &
\widetilde{\alpha}_{t,v} &= \sum_{d=1}^D \sum_{n=1}^{N_d} \bbI[w_{n,d}=v] \bbE[z_{n,d}=t] + \alpha_v. 
\end{align*}

The data local variables,~$z_{n,d}$,~$\bpsi_d$, and~$\bomega_d$, are conditionally independent across documents. Moreover, since the model is fully conjugate, the expectations required to update the global variables,~$\bbeta_t$,~$\bmu_\theta$, and~$\bSigma_\theta$ depend on sufficient statistics that are derived from summmations over documents.  Rather than summing over the entire corpus of documents, we can get an unbiased estimate of the sufficient statistics by considering a random subset, or mini-batch, per iteration. This is the key to stochastic variational inference (SVI) algorithms \cite{hoffman2013stochastic}, which have been widely successful in scalable topic modeling applications. Those same gains in scalability are readily applicable in this correlated topic model formulation.

\end{document}